# Anytime Induction of Low-cost, Low-error Classifiers: a Sampling-based Approach

**Saher Esmeir**                                ESAHER@CS.TECHNION.AC.IL
**Shaul Markovitch**                            SHAULM@CS.TECHNION.AC.IL
*Computer Science Department*
*Technion–Israel Institute of Technology*
*Haifa 32000, Israel*

## Abstract

Machine learning techniques are gaining prevalence in the production of a wide range of classifiers for complex real-world applications with nonuniform testing and misclassification costs. The increasing complexity of these applications poses a real challenge to resource management during learning and classification. In this work we introduce ACT (anytime cost-sensitive tree learner), a novel framework for operating in such complex environments. ACT is an anytime algorithm that allows learning time to be increased in return for lower classification costs. It builds a tree top-down and exploits additional time resources to obtain better estimations for the utility of the different candidate splits. Using sampling techniques, ACT approximates the cost of the subtree under each candidate split and favors the one with a minimal cost. As a stochastic algorithm, ACT is expected to be able to escape local minima, into which greedy methods may be trapped. Experiments with a variety of datasets were conducted to compare ACT to the state-of-the-art cost-sensitive tree learners. The results show that for the majority of domains ACT produces significantly less costly trees. ACT also exhibits good anytime behavior with diminishing returns.

## 1. Introduction

Traditionally, machine learning algorithms have focused on the induction of models with low expected error. In many real-word applications, however, several additional constraints should be considered. Assume, for example, that a medical center has decided to use machine learning techniques to build a diagnostic tool for heart disease. The comprehensibility of decision tree models (Hastie, Tibshirani, & Friedman, 2001, chap. 9) makes them the preferred choice on which to base this tool. Figure 1 shows three possible trees. The first tree (upper-left) makes decisions using only the results of cardiac catheterization (heart cath). This tree is expected to be highly accurate. Nevertheless, the high costs and risks associated with the heart cath procedure make this decision tree impractical. The second tree (lower-left) dispenses with the need for cardiac catheterization and reaches a decision based on a single, simple, inexpensive test: whether or not the patient complains of chest pain. Such a tree would be highly accurate: most people do not experience chest pain and are indeed healthy. The tree, however, does not distinguish between the costs of different types of errors. While a false positive prediction might result in extra treatments, a false negative prediction might put a person's life at risk. Therefore, a third tree (right) is preferred, one that attempts to minimize test costs and misclassification costs simultaneously.

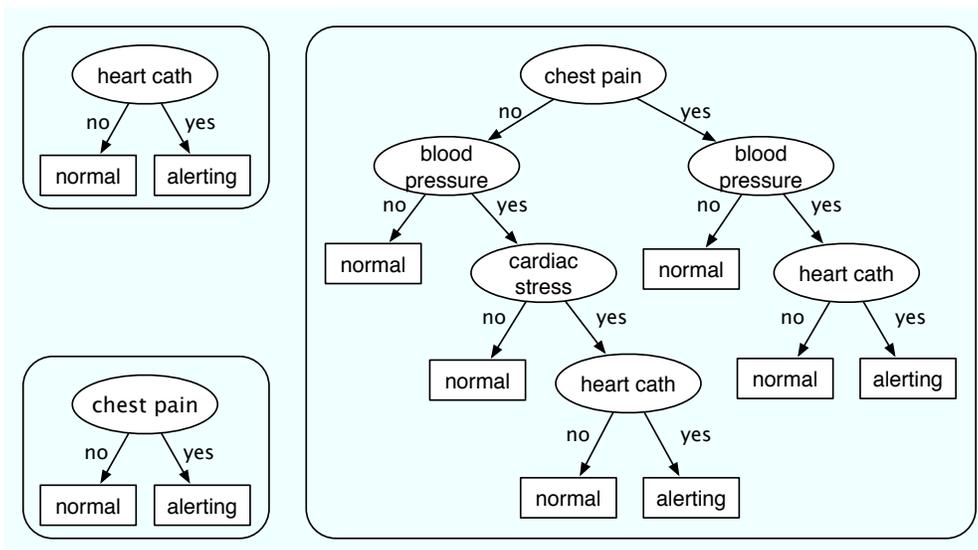

Figure 1: Three possible decision trees for diagnosis of heart diseases. The upper-left tree bases its decision solely on heart cath and is therefore accurate but prohibitively expensive. The lower-left tree dispenses with the need for heart cath and reaches a decision using a single, simple, and inexpensive test: whether or not the patient complains of chest pain. Such a tree would be highly accurate but does not distinguish between the costs of the different error types. The third (right-hand) tree is preferable: it attempts to minimize test costs and misclassification costs simultaneously.

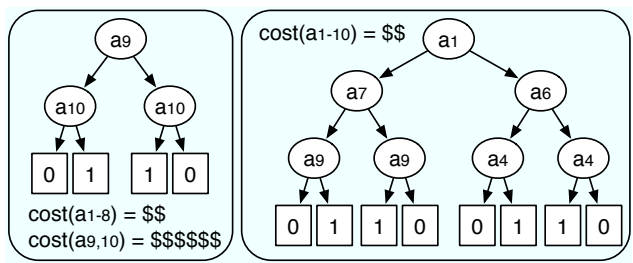

Figure 2: Left: an example of a difficulty greedy learners might face. Right: an example of the importance of context-based feature evaluation.

Finding a tree with the lowest expected *total cost* is at least NP-complete.[1] As in the cost insensitive case, a greedy heuristic can be used to bias the search towards low-cost trees. Decision Trees with Minimal Cost (DTMC), a greedy method that attempts to minimize both types of costs simultaneously, has been recently introduced (Ling, Yang, Wang, & Zhang, 2004; Sheng, Ling, Ni, & Zhang, 2006). A tree is built top-down, and a greedy split criterion that takes into account both testing and misclassification costs is used. The basic idea is to estimate the immediate reduction in total cost after each split, and to prefer the split with the maximal reduction. If no split reduces the cost on the training data, the induction process is stopped.

---

1. Finding the smallest consistent tree, which is an easier problem, is NP-complete (Hyafil & Rivest, 1976).

Although efficient, the DTMC approach can be trapped into a local minimum and produce trees that are not globally optimal. For example, consider the concept and costs described in Figure 2 (left). There are 10 attributes, of which only $a_9$ and $a_{10}$ are relevant. The cost of $a_9$ and $a_{10}$, however, is significantly higher than the others. Such high costs may hide the usefulness of $a_9$ and $a_{10}$, and mislead the learner into repeatedly splitting on $a_{1-8}$, which would result in a large, expensive tree. The problem would be intensified if $a_9$ and $a_{10}$ were interdependent, with a low immediate information gain (e.g., $a_9 \oplus a_{10}$). In that case, even if the costs were uniform, a local measure might fail to recognize the relevance of $a_9$ and $a_{10}$.

DTMC is appealing when learning resources are very limited. However, it requires a fixed runtime and cannot exploit additional resources to escape local minima. In many real-life applications, we are willing to wait longer if a better tree can be induced (Esmeir & Markovitch, 2006). For example, the importance of the model in saving patients' lives may convince the medical center to allocate 1 month to learn it. Algorithms that can exploit additional time to produce better solutions are called anytime algorithms (Boddy & Dean, 1994).

The ICET algorithm (Turney, 1995) was a pioneer in non-greedy search for a tree that minimizes test and misclassification costs. ICET uses genetic search to produce a new set of costs that reflects both the original costs and the contribution of each attribute in reducing misclassification costs. Then it builds a tree using the EG2 algorithm (Nunez, 1991) but with the evolved costs instead of the original ones. EG2 is a greedy cost-sensitive algorithm that builds a tree top-down and evaluates candidate splits by considering both the information gain they yield and their measurement costs. It does not, however, take into account the misclassification cost of the problem.

ICET was shown to significantly outperform greedy tree learners, producing trees of lower total cost. ICET can use additional time resources to produce more generations and hence widen its search in the space of costs. Because the genetic operations are randomized, ICET is more likely to escape local minima – into which EG2 with the original costs might be trapped. Nevertheless, two shortcomings limit ICET's ability to benefit from extra time. First, after the search phase, it uses the greedy EG2 algorithm to build the final tree. But because EG2 prefers attributes with high information gain (and low test cost), the usefulness of highly relevant attributes may be underestimated by the greedy measure in the case of hard-to-learn concepts where attribute interdependency is hidden. This will result in more expensive trees. Second, even if ICET overcomes the above problem by randomly reweighting the attributes, it searches the space of parameters globally, regardless of the context in the tree. This imposes a problem if an attribute is important in one subtree but useless in another. To better understand these shortcomings, consider the concept described by the tree in Figure 2 (right). There are 10 attributes with similar costs. The value of $a_1$ determines whether the target concept is $a_7 \oplus a_9$ or $a_4 \oplus a_6$. The interdependencies result in a low gain for all attributes. Because ICET assigns costs globally, the attributes will have similar costs as well. Therefore, ICET will not be able to recognize which one is relevant in which context. If the irrelevant attributes are cheaper, the problem is intensified and the model might end up relying on irrelevant attributes.

Recently, we have introduced the cost-insensitive LSID3 algorithm, which can induce more accurate trees when allocated more time (Esmeir & Markovitch, 2007a). The algorithm evaluates a candidate split by estimating the size of the smallest consistent tree under it. The estimation is based on sampling the space of consistent trees, where the size of the sample is determined in advance according to the allocated time. LSID3 is not designed, however, to minimize test and misclassification costs. In this work we build on LSID3 and propose $ACT$, an anytime cost-sensitive tree learner that can exploit additional time to produce lower-cost trees. Applying the sampling mechanism in the cost-sensitive setup, however, is not trivial and imposes three major challenges: (1) how to produce the sample, (2) how to evaluate the sampled trees, and (3) how to prune the induced trees. In Section 3 we show how these obstacles may be overcome.

In Section 4 we report an extensive set of experiments that compares ACT to several decision tree learners using a variety of datasets with costs assigned by human experts or automatically. The results show that ACT is significantly better for the majority of problems. In addition, ACT is shown to exhibit good anytime behavior with diminishing returns.

## 2. Cost-Sensitive Classification

Offline concept learning consists of two stages: the learning stage, where a set of labeled examples is used to induce a classifier; and the classification stage, where the induced classifier is used to classify unlabeled instances. These two stages involve different types of costs (Turney, 2000). Our primary goal in this work is to trade learning speed for a reduction in test and misclassification costs. To make the problem well defined, we need to specify: (1) how misclassification costs are represented, (2) how test costs are calculated, and (3) how we should combine both types of cost.

To answer these questions, we adopt the model described by Turney (1995). In a problem with $|C|$ different classes, a misclassification cost matrix $M$ is a $|C| \times |C|$ matrix whose $M_{i,j}$ entry defines the penalty of assigning the class $c_i$ to an instance that actually belongs to the class $c_j$. Typically, entries on the main diagonal of a classification cost matrix (no error) are all zero.

When classifying an example $e$ using a tree $T$, we propagate $e$ down the tree along a single path from the root of $T$ to one of its leaves. Let $\Theta(T, e)$ be the set of tests along this path. We denote by $cost(\theta)$ the cost of administering the test $\theta$. The testing cost of $e$ in $T$ is therefore $tcost(T, e) = \sum_{\theta \in \Theta} cost(\theta)$. Note that we use sets notation because tests that appear several times are charged for only once. In addition, the model described by Turney (1995) handles two special test types, namely *grouped* and *delayed* tests.

**Grouped Tests.** Some tests share a common cost, for which we would like to charge only once. Typically, the test also has an extra (possibly different) cost. For example, consider a tree path with tests like cholesterol level and glucose level. For both values to be measured, a blood test is needed. Taking blood samples to measure the cholesterol level clearly lowers the cost of measuring the glucose level. Formally, each test possibly belongs to a group.[2] If it's the first test from the group to be administered, we charge for the full cost. If another test from the same group has already been administered earlier in the decision path, we charge only for the marginal cost.

**Delayed Tests.** Sometimes the outcome of a test cannot be obtained immediately, e.g., lab test results. Such tests, called *delayed* tests, force us to wait until the outcome is available. Alternatively, Turney (1995) suggests taking into account all possible outcomes: when a delayed test is encountered, all the tests in the subtree under it are administered and charged for. Once the result of the delayed test is available, the prediction is at hand. One problem with this setup is that it follows all paths in the subtree, regardless of the outcome of non-delayed costs. Moreover, it is not possible to distinguish between the delays different tests impose: for example, one result might be ready after several minutes while another only after a few days. In this work we do not handle delayed tests, but we do explain how ACT can be modified to take them into account.

After the test and misclassification costs have been measured, an important question remains: How should we combine them? Following Turney (1995), we assume that both cost types are given in the same scale. A more general model would require a utility function that combines both types. Qin, Zhang, and Zhang (2004) presented a method to handle the two kinds of cost scales by setting a maximal budget for one kind of cost and minimizing the other one. Alternatively, patient preferences can be elicited and summarized as a utility function (Lenert & Soetikno, 1997).

Note that the algorithm we introduce in this paper can be adapted to any cost model. An important property of our cost-sensitive setup is that maximizing generalization accuracy, which is the goal of most existing learners, can be viewed as a special case: when accuracy is the only objective, test costs are ignored and misclassification cost is uniform.

## 3. The ACT Algorithm

ACT, our proposed anytime framework for induction of cost-sensitive decision trees, builds on the recently introduced LSID3 algorithm. LSID3 adopts the general top-down scheme for induction of decision trees (TDIDT): it starts from the entire set of training examples, partitions it into subsets by testing the value of an attribute, and then recursively builds subtrees. Unlike greedy inducers, LSID3 invests more time resources for making better split decisions. For every candidate split, LSID3 attempts to estimate the size of the resulting subtree were the split to take place. Following Occam's razor (Blumer, Ehrenfeucht, Haussler, & Warmuth, 1987; Esmeir & Markovitch, 2007b), it favors the one with the smallest expected size.

The estimation is based on a biased sample of the space of trees rooted at the evaluated attribute. The sample is obtained using a stochastic version of ID3 (Quinlan, 1986), which we call SID3. In SID3, rather than choosing an attribute that maximizes the information gain $\Delta I$ (as in ID3), we choose the splitting attribute semi-randomly. The likelihood that an attribute will be chosen is proportional to its information gain. Due to its randomization,

---

2. In this model each test may belong to a single group. However, it is easy to extend our work to allow tests that belong to several groups.

---

**Procedure** LSID3-Choose-Attribute($E, A, r$)
   **If** $r = 0$
      **Return** ID3-Choose-Attribute(E, A)
   **Foreach** $a \in A$
      **Foreach** $v_i \in \mathrm{domain}(a)$
        $E_i \leftarrow \{e \in E \mid a(e) = v_i\}$
        $min_i \leftarrow \infty$
        **Repeat** $r$ **times**
          $T \leftarrow \mathrm{SID3}(E_i, A - \{a\})$
          $min_i \leftarrow \min{(min_i, \mathrm{Size}(T))}$
      $total_a \leftarrow \sum_{i=1}^{|\mathrm{domain}(a)|} min_i$
   **Return** $a$ for which $total_a$ is minimal

---

Figure 3: Attribute selection in LSID3

repeated invocations of SID3 result in different trees. For each candidate attribute $a$, LSID3 invokes SID3 $r$ times to form a sample of $r$ trees rooted at $a$, and uses the size of the smallest tree in the sample to evaluate $a$. Obviously, when $r$ is larger, the resulting size estimations are expected to be more accurate, improving the final tree. Consider, for example, a 3-XOR concept with several additional irrelevant attributes. For LSID3 to prefer one of the relevant attributes at the root, one of the trees in the samples of the relevant attributes must be the smallest. The probability for this event increases with the increase in sample size.

LSID3 is a contract anytime algorithm parameterized by $r$, the sample size. Additional time resources can be utilized by forming larger samples. Figure 3 lists the procedure for attribute selection as applied by LSID3. Let $m = |E|$ be the number of examples and $n = |A|$ be the number of attributes. The runtime complexity of LSID3 is $O(rmn^3)$. LSID3 was shown to exhibit good anytime behavior with diminishing returns. When applied to hard concepts, it produced significantly better trees than ID3 and C4.5.

ACT takes the same sampling approach as LSID3. The three major components of LSID3 that need to be replaced in order to adapt it for cost-sensitive problems are: (1) sampling the space of trees, (2) evaluating a tree, and (3) pruning a tree.

### 3.1 Obtaining the Sample

LISD3 uses SID3 to bias the samples towards small trees. In ACT, however, we would like to bias our sample towards low-cost trees. For this purpose, we designed a stochastic version of the EG2 algorithm, which attempts to build low cost trees greedily. In EG2, a tree is built top-down, and the test that maximizes ICF is chosen for splitting a node, where,

$$\mathrm{ICF}\left(\theta\right) = \frac{2^{\Delta I(\theta)} - 1}{\left(\mathrm{cost}\left(\theta\right) + 1\right)^{w}}.$$

$\Delta I$ is the information gain (as in ID3). The parameter $w \in [0, 1]$ controls the bias towards lower cost attributes. When $w = 0$, test costs are ignored and ICF relies solely

---

**Procedure** SEG2-Choose-Attribute$(E, A)$
    **Foreach** $a \in A$
        $\Delta I(a) \leftarrow$ Information-Gain$(E, a)$
        $c(a) \leftarrow$ Cost$(a)$
        $p(a) \leftarrow \frac{2^{\Delta I(a)} - 1}{(c(a)+1)^w}$
    $a^* \leftarrow$ Choose attribute at random from $A$;
        for each attribute $a$, the probability
        of selecting it is proportional to $p(a)$
    **Return** $a^*$

---

Figure 4: Attribute selection in SEG2

on the information gain. Larger values of $w$ strengthen the effect of test costs on ICF. We discuss setting the value of $w$ in Section 3.5.

In stochastic EG2 (SEG2), we choose splitting attributes semi-randomly, proportionally to their ICF. Because SEG2 is stochastic, we expect to be able to escape local minima for at least some of the trees in the sample. Figure 4 formalizes the attribute selection component in SEG2. To obtain a sample of size $r$, ACT uses EG2 once and SEG2 $r - 1$ times. EG2 and SEG2 are given direct access to context-based costs, i.e., if an attribute has already been tested, its cost is zero and if another attribute that belongs to the same group has been tested, a group discount is applied.

## 3.2 Evaluating a Subtree

LSID3 is a cost-insensitive learning algorithm. As such, its main goal is to maximize the expected accuracy of the learned tree. Occam's razor states that given two consistent hypotheses, the smaller one is likely to be more accurate. Following Occam's razor, LSID3 uses the tree size as a preference bias and favors splits that are expected to reduce its final size.

In a cost-sensitive setup, however, our goal is to minimize the expected *total* cost of classification. Therefore, rather than choosing an attribute that minimizes the size, we would like to choose one that minimizes the total cost. Given a decision tree, we need to come up with a procedure that estimates the expected cost of using the tree to classify a future case. This cost has two components: the test cost and the misclassification cost.

### 3.2.1 Estimating Test Costs

Assuming that the distribution of future cases would be similar to that of the learning examples, we can estimate the test costs using the training data. Given a tree, we calculate the average test cost of the training examples and use it to estimate the test cost of new cases. For a (sub)tree $T$ built from $E$, a set of $m$ training examples, we denote the average cost of traversing $T$ for an example from $E$ by

$$\overline{tcost}(T, E) = \frac{1}{m} \sum_{e \in E} tcost(T, e).$$

The estimated test cost for an unseen example $e^*$ is therefore $\widehat{tcost}(T, e^*) = \overline{tcost}(T, E)$.

Observe that costs are calculated in the relevant context. If an attribute $a$ has already been tested in upper nodes, we will not charge for testing it again. Similarly, if an attribute from a group $g$ has already been tested, we will apply a group discount to the other attributes from $g$. If a delayed attribute is encountered, we sum the cost of the entire subtree.

### 3.2.2 Estimating Misclassification Costs

How to go about estimating the cost of misclassification is not obvious. The tree size can no longer be used as a heuristic for predictive errors. Occam's razor allows the comparison of two consistent trees but provides no means for estimating accuracy. Moreover, tree size is measured in a different currency than accuracy and hence cannot be easily incorporated in the cost function.

Rather than using the tree size, we propose a different estimator: the expected error (Quinlan, 1993). For a leaf with $m$ training examples, of which $s$ are misclassified, the expected error is defined as the upper limit on the probability for error, i.e., $EE(m, s, cf) = U_{cf}^{bin}(m, s)$, where $cf$ is the confidence level and $U^{bin}$ is the upper limit of the confidence interval for binomial distribution. The expected error of a tree is the sum of the expected errors in its leaves.

Originally, the expected error was used by C4.5's *error-based pruning* to predict whether a subtree performs better than a leaf. Although lacking a theoretical basis, it was shown experimentally to be a good heuristic. In ACT we use the expected error to approximate the misclassification cost. Assume a problem with $|C|$ classes and a misclassification cost matrix $M$. Let $c$ be the class label in a leaf $l$. Let $m_l$ be the total number of examples in $l$ and $m_l^i$ be the number of examples in $l$ that belong to class $i$. When the penalties for predictive errors are uniform ($M_{i,j} = mc$), the estimated misclassification cost in $l$ is

$$\widehat{mcost}(l) = \text{EE}(m_l, m_l - m_l^c, cf) \cdot mc.$$

In a problem with nonuniform misclassification costs, $mc$ should be replaced by the cost of the actual errors the leaf is expected to make. These errors are obviously unknown to the learner. One solution is to estimate each error type separately using confidence intervals for multinomial distribution and multiply it by the associated cost:

$$\widehat{mcost}(l) = \sum_{i \neq c} U_{cf}^{mul}(m_l, m_l^i, |C|) \cdot mc.$$

Such approach, however, would result in an overly pessimistic approximation, mainly when there are many classes. Alternatively, we compute the expected error as in the uniform case and propose replacing $mc$ by the weighted average of the penalty for classifying an instance as $c$ while it belongs to another class. The weights are derived from the proportions $\frac{m_l^i}{m_l - m_l^c}$ using a generalization of Laplace's law of succession (Good, 1965, chap. 4):

$$\widehat{mcost}(l) = \text{EE}(m_l, m_l - m_l^c, cf) \cdot \sum_{i \neq c} \left( \frac{m_l^i + 1}{m_l - m_l^c + |C| - 1} \cdot M_{c,i} \right).$$

Note that in a problem with $C$ classes, the average is over $C - 1$ possible penalties because $M_{c,c} = 0$. Hence, in a problem with two classes $c_1, c_2$ if a leaf is marked as $c_1$, $mc$

---

**Procedure** ACT-Choose-Attribute$(E, A, r)$
    **If** $r = 0$
        **Return** EG2-Choose-Attribute(E, A)
    **Foreach** $a \in A$
        **Foreach** $v_i \in \mathrm{domain}(a)$
            $E_i \leftarrow \{e \in E \mid a(e) = v_i\}$
            $T \leftarrow \mathrm{EG2}(a, E_i, A - \{a\})$
            $min_i \leftarrow$ Total-Cost$(T, E_i)$
            **Repeat** $r - 1$ **times**
                $T \leftarrow \mathrm{SEG2}(a, E_i, A - \{a\})$
                $min_i \leftarrow \min\left(min_i, \text{Total-Cost}(T, E_i)\right)$
        $total_a \leftarrow \text{Cost}(a) + \sum_{i=1}^{|\mathrm{domain}(a)|} min_i$
    **Return** $a$ for which $total_a$ is minimal

Figure 5: Attribute selection in ACT

would be replaced by $M_{1,2}$. When classifying a new instance, the expected misclassification cost of a tree $T$ built from $m$ examples is the sum of the expected misclassification costs in the leaves divided by $m$:

$$\widehat{mcost}(T) = \frac{1}{m} \sum_{l \in L} \widehat{mcost}(l),$$

where $L$ is the set of leaves in $T$. Hence, the expected total cost of $T$ when classifying a single instance is:

$$\widehat{total}(T, E) = \widehat{tcost}(T, E) + \widehat{mcost}(T).$$

An alternative approach that we intend to explore in future work is to estimate the cost of the sampled trees using the cost for a set-aside validation set. This approach is attractive mainly when the training set is large and one can afford setting aside a significant part of it.

### 3.3 Choosing a Split

Having decided about the sampler and the tree utility function, we are ready to formalize the tree growing phase in ACT. A tree is built top-down. The procedure for selecting a splitting test at each node is listed in Figure 5 and illustrated in Figure 6. We give a detailed example of how ACT chooses splits and explain how the split selection procedure is modified for numeric attributes.

#### 3.3.1 Choosing a Split: Illustrative Examples

ACT's evaluation is cost-senstive both in that it considers test and error costs simultaneously and in that it can take into account different error penalties. To illustrate this let us consider a two-class problem with $mc = 100\$$ (uniform) and 6 attributes, $a_1, \ldots, a_6$, whose costs are $10\$$. The training data contains 400 examples, out of which 200 are positive and 200 are negative.

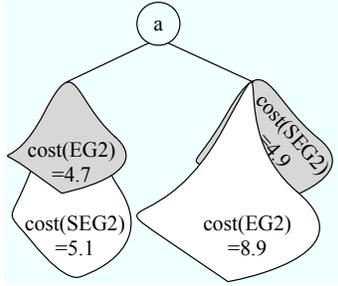

Figure 6: Attribute evaluation in ACT. Assume that the cost of $a$ in the current context is 0.4. The estimated cost of a subtree rooted at $a$ is therefore $0.4 + min(4.7, 5.1) + min(8.9, 4.9) = 10$.

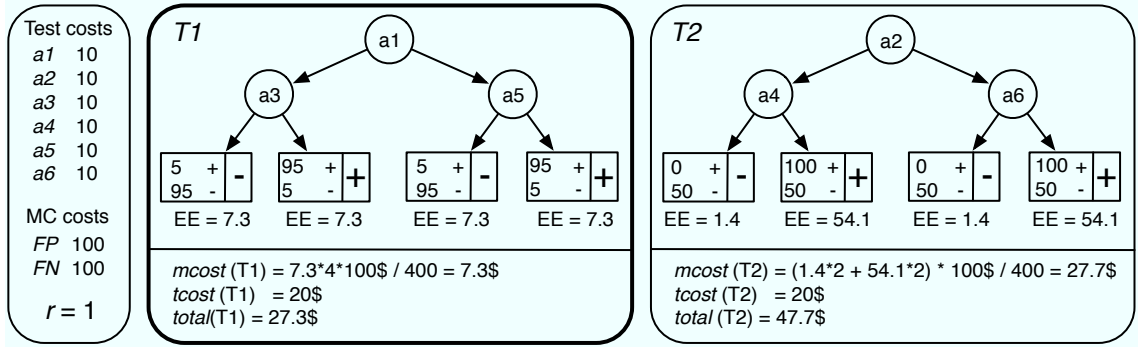

Figure 7: Evaluation of tree samples in ACT. The leftmost column defines the costs: 6 attributes with identical cost and uniform error penalties. $T1$ was sampled for $a_1$ and $T2$ for $a_2$. $EE$ stands for the expected error. Because the total cost of $T_1$ is lower, ACT would prefer to split on $a_1$.

Assume that we have to choose between $a_1$ and $a_2$, and that $r = 1$. Let the trees in Figure 7, denoted $T1$ and $T2$, be those sampled for $a_1$ and $a_2$ respectively. The expected error costs of $T_1$ and $T_2$ are:[3]

$$\widehat{mcost}(T_1) = \frac{1}{400}\left(4 \cdot \text{EE}\,(100, 5, 0.25)\right) \cdot 100\$ = \frac{4 \cdot 7.3}{400} \cdot 100\$ = 7.3\$$$

$$\widehat{mcost}(T_2) = \frac{1}{400}\left(2 \cdot \text{EE}\,(50, 0, 0.25) \cdot 100\$ + 2 \cdot \text{EE}\,(150, 50, 0.25) \cdot 100\$\right)$$

$$= \frac{2 \cdot 1.4 + 2 \cdot 54.1}{400} \cdot 100\$ = 27.7\$$$

When both test and error costs are involved, ACT considers their sum. Since the test cost of both trees is identical (20$), ACT would prefer to split on $a_1$. If, however, the cost

---

3. In this example we set $cf$ to 0.25, as in C4.5. In Section 3.5 we discuss how to tune $cf$.

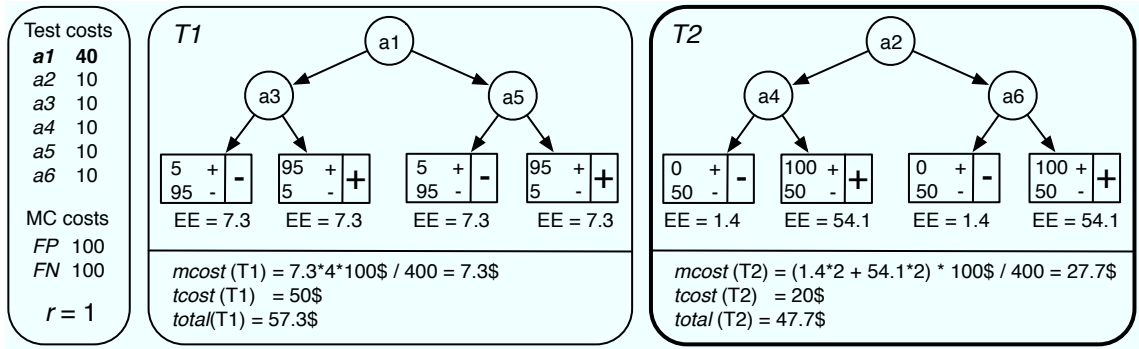

Figure 8: Evaluation of tree samples in ACT. The leftmost column defines the costs: 6 attributes with identical cost (except for the expensive $a_1$) and uniform error penalties. $T1$ was sampled for $a_1$ and $T2$ for $a_2$. Because the total cost of $T_2$ is lower, ACT would prefer to split on $a_2$.

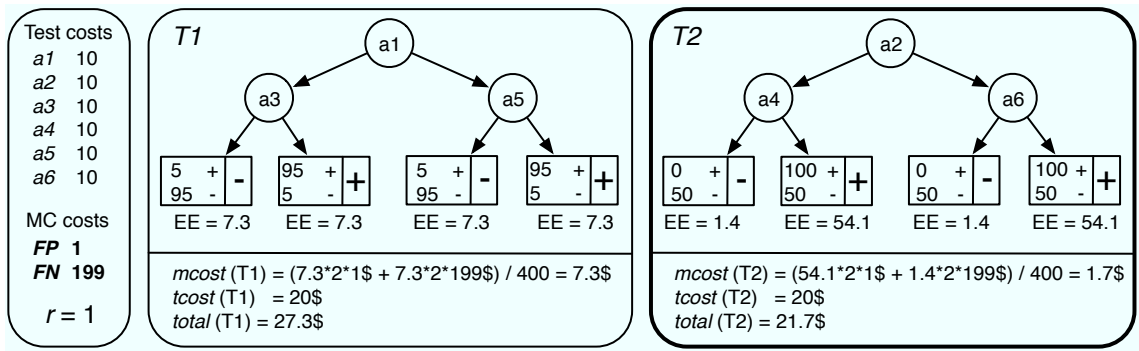

Figure 9: Evaluation of tree samples in ACT. The leftmost column defines the costs: 6 attributes with identical cost and nonuniform error penalties. $T1$ was sampled for $a_1$ and $T2$ for $a_2$. Because the total cost of $T_2$ is lower, ACT would prefer to split on $a_2$.

of $a_1$ were 40\$, as in Figure 8, $tcost(T1)$ would become 50\$ and the total cost of $T1$ would become 57.3\$, while that of $T2$ would remain 47.7\$. Hence, in this case ACT would split on $a_2$.

To illustrate how ACT handles nonuniform error penalties, let us assume that the cost of all attributes is again 10\$, while the cost of a false positive ($FP$) is 1\$ and the cost of a false negative ($FN$) is 199\$. Let the trees in Figure 9, denoted $T1$ and $T2$, be those sampled for $a_1$ and $a_2$ respectively. As in the first example, only misclassification costs play a role because the test costs of both trees is the same. Although on average the misclassification cost is also 100\$, ACT now evaluates these trees differently:

$$\widehat{mcost}(T_1) = \frac{1}{400}\left(2 \cdot \text{EE}\left(100, 5, 0.25\right) \cdot 1\$ + 2 \cdot \text{EE}\left(100, 5, 0.25\right) \cdot 199\$\right)$$

$$= \frac{2 \cdot 7.3 \cdot 1\$ + 2 \cdot 7.3.1 \cdot 199\$}{400} = 7.3\$$$

$$\widehat{mcost}(T_2) = \frac{1}{400}\left(2 \cdot \text{EE}\left(50, 0, 0.25\right) \cdot 199\$ + 2 \cdot \text{EE}\left(100, 50, 0.25\right) \cdot 1\$\right)$$

$$= \frac{2 \cdot 1.4 \cdot 199\$ + 2 \cdot 54.1 \cdot 1\$}{400} = 1.7\$$$

Therefore, in the nonuniform setup, ACT would prefer $a_2$. This makes sense because in the given setup we prefer trees that may result in more false positives but reduce the number of expensive false negatives.

### 3.3.2 Choosing a Split when Attributes are Numeric

The selection procedure as formalized in Figure 5 must be modified slightly when an attribute is numeric: rather than iterating over the values the attribute can take, we first pick $r$ tests (split points) with the highest information gain and then invoke EG2 once for each split point. This guarantees that numeric and nominal attributes get the same resources. Chickering, Meek, and Rounthwaite (2001) introduced several techniques for generating a small number of candidate split points dynamically with little overhead. In the future we intend to apply these techniques to select $r$ points, each of which will be evaluated with a single invocation of EG2.

## 3.4 Cost-Sensitive Pruning

Pruning plays an important role in decision tree induction. In cost-insensitive environments, the main goal of pruning is to simplify the tree in order to avoid overfitting the training data. A subtree is pruned if the resulting tree is expected to yield a lower error.

When test costs are taken into account, pruning has another important role: reducing test costs in a tree. Keeping a subtree is worthwhile only if its expected reduction in misclassification costs is larger than the cost of the tests in that subtree. If the misclassification cost is zero, it makes no sense to keep any split in the tree. If, on the other hand, the misclassification cost is much larger than the test costs, we would expect similar behavior to the cost-insensitive setup.

To handle this challenge, we propose a novel approach for cost-sensitive pruning. As in error-based pruning (Quinlan, 1993), we scan the tree bottom-up. Then we compare the expected total cost of each subtree to that of a leaf. If a leaf is expected to perform better, the subtree is pruned.

The cost of a subtree is estimated as described in Section 3.2. Formally, let $E$ be the set of training examples that reach a subtree $T$, and let $m$ be the size of $E$. Assume that $s$ examples in $E$ do not belong to the default class.[4] Let $L$ be the set of leaves in $T$. We

---

4. If misclassification costs are uniform, the default class is the majority class. Otherwise, it is the class that minimizes the misclassification cost in the node.

prune $T$ into a leaf if:

$$\frac{1}{m} \cdot EE(m, s, cf) \cdot mc \leq \widehat{tcost}(T, E) + \sum_{l \in L} \widehat{mcost}(l).$$

The above assumes a uniform misclassification cost $mc$. In the case of nonuniform penalties, we multiply the expected error by the average misclassification cost.

An alternative approach for post-pruning is early stopping of the growing phase. For example, one could limit the depth of the tree, require a minimal number of examples in each child (as in C4.5), or prevent splitting nodes when the splitting criterion fails to exceed a predetermined threshold (as in DTMC). Obviously, any pre-pruning condition can also be applied as part of the post-pruning procedure. The advantage of post-pruning, however, is its ability to estimate the effect of a split on the entire subtree below it, and not only on its immediate successors (horizon effect).

Consider for example the 2-XOR problem $a \oplus b$. Splitting on neither $a$ nor $b$ would have a positive gain and hence the growing would be stopped. If no pre-pruning is allowed, the optimal tree would be found and would not be post-pruned because the utility of the splits is correctly measured. Frank (2000) reports a comprehensive study about pruning of decision trees, in which he compared pre- to post-pruning empirically in a cost-insensitive setup. His findings show that the advantage of post-pruning on a variety of UCI datasets is not significant. Because pre-pruning is computationally more efficient, Frank concluded that, in practice, it might be a viable alternative to post-pruning. Despite these results, we decided to use post-pruning in ACT, for the following reasons:

1. Several concepts not represented in the UCI repository may appear in real-world problems. For example, parity functions naturally arise in real-world problems, such as the Drosophila survival concept (Page & Ray, 2003).

2. When costs are involved, the horizon effect may appear more frequently because high costs may hide good splits.

3. In our anytime setup the user is willing to wait longer in order to obtain a good tree. Since post-pruning takes even less time than the induction of a single greedy tree, the extra cost of post-pruning is minor.

In the future we plan to add a pre-pruning parameter which will allow early stopping when resources are limited. Another interesting direction for future work would be to post-prune the final tree but pre-prune the lookahead trees that form the samples. This would reduce the runtime at the cost of less accurate estimations for the utility of each candidate split.

## 3.5 Setting the Parameters of ACT

In addition to $r$, the sample size, ACT is parameterized by $w$, which controls the weight of the test costs in EG2, and $cf$, the confidence factor used both for pruning and for error estimation. ICET tunes $w$ and $cf$ using genetic search. In ACT we considered three different alternatives: (1) keeping EG2's and C4.5's default values $w = 1$ and $cf = 0.25$, (2) tuning the values using cross-validation, and (3) setting the values a priori, as a function of the problem costs.

While the first solution is the simplest, it does not exploit the potential of adapting the sampling mechanism to the specific problem costs. Although tuning the values using grid search would achieve good results, it may be costly in terms of runtime. For example, if we had 5 values for each parameter and used 5-fold cross-validation, we would need to run ACT 125 times for the sake of tuning alone. In our anytime setup this time could be invested to invoke ACT with a larger $r$ and hence improve the results. Furthermore, the algorithm would not be able to output any valid solution before the tuning stage is finished. Alternatively, we could try to tune the parameters by invoking the much faster EG2, but the results would not be as good because the optimal values for EG2 are not necessarily good for ACT.

The third approach, which we chose for our experiments, is to set $w$ and $cf$ in advance, according to the problem specific costs. $w$ is set inverse proportionally to the misclassification cost: a high misclassification cost results in a smaller $w$, reducing the effect of attribute costs on the split selection measure. The exact formula is:

$$w = 0.5 + e^{-x},$$

where $x$ is the average misclassification cost (over all non-diagnoal entries in $M$) divided by $TC$, the cost if we take all tests. Formally,

$$x = \frac{\sum_{i \neq j} M_{i,j}}{(|C| - 1) \cdot |C| \cdot TC}.$$

In C4.5 the default value of $cf$ is 0.25. Larger $cf$ values result in less pruning. Smaller $cf$ values lead to more aggressive pruning. Therefore, in ACT we set $cf$ to a value in the range $[0.2, 0.3]$; the exact value depends on the problem cost. When test costs are dominant, we prefer aggressive pruning and hence a low value for $cf$. When test costs are negligible, we prefer to prune less. The same value of $cf$ is also used to estimate the expected error. Again, when test costs are dominant, we can afford a pessimistic estimate of the error, but when misclassification costs are dominant, we would prefer that the estimate be closer to the error rate in the training data. The exact formula for setting $cf$ is:

$$cf = 0.2 + 0.05(1 + \frac{x-1}{x+1}).$$

## 4. Empirical Evaluation

We conducted a variety of experiments to test the performance and behavior of ACT. First we introduce a novel method for automatic adaption of existing datasets to the cost-sensitive setup. We then describe our experimental methodology and its motivation. Finally we present and discuss our results.

### 4.1 Datasets

Typically, machine learning researchers use datasets from the UCI repository (Asuncion & Newman, 2007). Only five UCI datasets, however, have assigned test costs.[5] We include these datasets in our experiments. Nevertheless, to gain a wider perspective, we have developed an automatic method that assigns costs to existing datasets. The method is parameterized with:

1. $cr$, the cost range.

2. $g$, the number of desired groups as a percentage of the number of attributes. In a problem with $|A|$ attributes, there are $g \cdot |A|$ groups. The probability for an attribute to belong to each of these groups is $\frac{1}{g \cdot |A| + 1}$, as is the probability for it not to belong to any of the groups.

3. $d$, the number of delayed tests as a percentage of the number of attributes.

4. $\varphi$, the group discount as a percentage of the minimal cost in the group (to ensure positive costs).

5. $\rho$, a binary flag which determines whether costs are drawn randomly, uniformly ($\rho = 0$) or semi-randomly ($\rho = 1$): the cost of a test is drawn proportionally to its information gain, simulating a common case where valuable features tend to have higher costs. In this case we assume that the cost comes from a truncated normal distribution, with the mean being proportional to the gain.

---

5. Costs for these datasets have been assigned by human experts (Turney, 1995).

Using this method, we assigned costs to 25 datasets: 20 arbitrarily chosen UCI datasets[6] and 5 datasets that represent hard concepts and have been used in previous research. Appendix A gives detailed descriptions of these datasets.

Due to the randomization in the cost assignment process, the same set of parameters defines an infinite space of possible costs. For each of the 25 datasets we sampled this space 4 times with

$$cr = [1, 100], g = 0.2, d = 0, \varphi = 0.8, \rho = 1.$$

These parameters were chosen in an attempt to assign costs in a manner similar to that in which real costs are assigned. In total, we have 105 datasets: 5 assigned by human experts and 100 with automatically generated costs.[7]

Cost-insensitive learning algorithms focus on accuracy and therefore are expected to perform well when testing costs are negligible relative to misclassification costs. However, when testing costs are significant, ignoring them would result in expensive classifiers. Therefore, evaluating cost-sensitive learners requires a wide spectrum of misclassification costs. For each problem out of the 105, we created 5 instances, with uniform misclassification costs $mc = 100, 500, 1000, 5000, 10000$. Later on, we also consider nonuniform misclassification costs.

## 4.2 Methodology

We start our experimental evaluation by comparing ACT, given a fixed resource allocation, with several other cost-sensitive and cost-insensitive algorithms. Next we compare the anytime behavior of ACT to that of ICET. Finally, we evaluate the algorithms with two modifications on the problem instances: random test cost assignment and nonuniform misclassification costs.

---

6. The chosen UCI datasets vary in size, type of attributes, and dimension.

7. The additional 100 datasets are available at http://www.cs.technion.ac.il/~esaher/publications/cost.

### 4.2.1 Compared Algorithms

ACT is compared to the following algorithms:

- *C4.5*: A cost-insensitive greedy decision tree learner. The algorithm has been re-implemented following the details in (Quinlan, 1993) and the default parameters have been used.

- *LSID3*: A cost-insensitive anytime decision tree learner. As such it uses extra time to induce trees of higher accuracy. It is not able, however, to exploit additional allotted time to reduce classification costs.

- *IDX*: A greedy top-down learner that prefers splits that maximize $\frac{\Delta I}{c}$ (Norton, 1989). The algorithm does not take into account misclassification costs. IDX has been implemented on top of C4.5, by modifying the split selection criteria.

- *CSID3*: A greedy top-down learner that prefers splits that maximize $\frac{\Delta I^2}{c}$ (Tan & Schlimmer, 1989). The algorithm does not take into account misclassification costs. CSID3 has been implemented on top of C4.5, by modifying the split selection criteria.

- *EG2*: A greedy top-down learner that prefers splits that maximize $\frac{2^{\Delta I(\theta)}-1}{(\text{cost}(\theta)+1)^w}$ (Nunez, 1991). The algorithm does not take into account misclassification costs. EG2 has been implemented on top of C4.5, by modifying the split selection criteria.

- *DTMC*: DTMC was implemented by following the original pseudo-code (Ling et al., 2004; Sheng et al., 2006). However, the original pseudo-code does not support continuous attributes and multiple class problems. We added support to continuous attributes, as in C4.5's dynamic binary-cut discretization, with the cost reduction replacing gain ratio for selecting cutting points. The extension to multiple class problems was straightforward. Note that DTMC does not post-prune the trees but only pre-prunes them.

- *ICET*: ICET has been reimplemented following the detailed description given by Turney (1995). To verify the results of the reimplementation, we compared them with those reported in the literature. We followed the same experimental setup and used the same 5 datasets. The results are indeed similar: the basic version of ICET achieved an average cost of 50.8 in our reimplementation vs. 50 reported originally. One possible reason for the slight difference may be that the initial population of the genetic algorithm is randomized, as are the genetic operators and the process of partitioning the data into training, validating, and testing sets. In his paper, Turney introduced a seeded version of ICET, which includes the true costs in the initial population, and reported it to perform better than the unseeded version. Therefore, we use the seeded version for our comparison. The other parameters of ICET are the default ones.

### 4.2.2 Normalized Cost

As Turney (1995) points out, using the average cost of classification for each dataset is problematic because: (1) the cost differences of the algorithms become relatively small as the misclassification cost increases, (2) it is difficult to combine the results for multiple datasets in a fair manner (e.g., average), and (3) it is difficult to combine the average of the different misclassification costs. To overcome these problems, Turney suggests normalizing the average cost of classification by dividing it by the *standard cost*. Let $TC$ be the cost if we take all tests. Let $f_i$ be the frequency of class $i$ in the data. The error if the response is always class $i$ is therefore $(1 - f_i)$. The standard cost is defined as

$$TC + min_i \left(1 - f_i\right) \cdot max_{i,j}\left(M_{i,j}\right).$$

The standard cost is an approximation for the maximal cost in a given problem. It consists of two components: the maximal test cost and the misclassification cost if the classifier achieves only the baseline accuracy (e.g., a majority-based classifier when error costs are uniform). Because some classifiers may perform even worse than the baseline accuracy, the standard cost is not strictly an upper bound on real cost. In most of our experiments, however, it has not been exceeded.

### 4.2.3 Statistical Significance

For each problem out of the 105, a single 10-fold cross-validation experiment was conducted. The same partition to train-test sets was used for all compared algorithms. To determine statistical significance of the performance differences between ACT, ICET, and DTMC we used two tests:

- Paired t-test with $\alpha = 5\%$ confidence. For each problem out of the 105 and for each pair of algorithms, we have 10 pairs of results obtained from the 10-fold cross validation runs. We used paired t-test to determine weather the difference between the two algorithms on a given problem is significant (rejecting the null hypothesis that the algorithms do not differ in their performance). Then, we count for each algorithm how many times it was a significant winner.

- Wilcoxon test (Demsar, 2006), which compares classifiers over multiple datasets and states whether one method is significantly better than the other ($\alpha = 5\%$).

## 4.3 Fixed-time Comparison

For each of the $105 \times 5$ problem instances, we ran the different algorithms, including ACT with $r = 5$. We chose $r = 5$ so the average runtime of ACT would be shorter than ICET over all problems. The other methods have much shorter runtime due to their greedy nature.

Table 1 summarizes the results.[8] Each pair of numbers represents the average normalized cost and its associated confidence interval ($\alpha = 5\%$). Figure 10 illustrates the average results and plots the normalized costs for the different algorithms and misclassification costs.

Statistical significance test results for ACT, ICET, and DTMC are given in Table 2. The algorithms are compared using both the t-test and the Wilcoxon test. The table lists the number of t-test wins for each algorithm out of the 105 datasets, as well as the winner, if any, when the Wilcoxon test was applied.

---

8. The full results are available at http://www.cs.technion.ac.il/∼esaher/publications/cost.

Table 1: Average cost of classification as a percentage of the standard cost of classification for different $mc$ values. The numbers represent the average over all 105 datasets and the associated confidence intervals ($\alpha = 5\%$).

| $mc$ | C4.5 | LSID3 | IDX | CSID3 | EG2 | DTMC | ICET | ACT |
|------|------|-------|-----|-------|-----|------|------|-----|
| 100 | 50.6 ±4.2 | 45.3 ±3.7 | 34.4 ±3.6 | 41.7 ±3.8 | 35.1 ±3.6 | 14.6 ±1.8 | 24.3 ±3.1 | 15.2 ±1.9 |
| 500 | 49.9 ±4.2 | 43.0 ±3.9 | 42.4 ±3.6 | 45.2 ±3.9 | 42.5 ±3.6 | 37.7 ±3.1 | 36.3 ±3.1 | 34.5 ±3.2 |
| 1000 | 50.4 ±4.6 | 42.4 ±4.5 | 47.5 ±4.2 | 47.8 ±4.4 | 47.3 ±4.3 | 47.1 ±3.8 | 40.6 ±3.9 | 39.1 ±4.2 |
| 5000 | 53.3 ±5.9 | 43.6 ±6.1 | 58.1 ±5.9 | 54.3 ±5.9 | 57.3 ±5.9 | 57.6 ±5.2 | 45.7 ±5.6 | 41.5 ±5.7 |
| 10000 | 54.5 ±6.4 | 44.5 ±6.6 | 60.8 ±6.4 | 56.2 ±6.4 | 59.9 ±6.4 | 59.5 ±5.6 | 47.1 ±6.0 | 41.4 ±6.0 |

Table 2: DTMC vs. ACT and ICET vs. ACT using statistical tests. For each $mc$, the first column lists the number of t-test significant wins while the second column gives the winner, if any, as implied by a Wilcoxon test over all datasets with $\alpha = 5\%$.

| | $t-test$ WINS | | | | $Wilcoxon$ WINNER | |
|------|------|------|------|------|------|------|
| $mc$ | DTMC | vs. | ACT | ICET | vs. ACT | DTMC vs. ACT | ICET vs. ACT |
| 100 | 14 | 3 | 4 | 54 | DTMC | ACT |
| 500 | 9 | 29 | 5 | 23 | ACT | ACT |
| 1000 | 7 | 45 | 12 | 24 | ACT | ACT |
| 5000 | 7 | 50 | 15 | 21 | ACT | ACT |
| 10000 | 6 | 56 | 7 | 24 | ACT | - |

When misclassification cost is relatively small ($mc = 100$), ACT clearly outperforms ICET, with 54 significant wins as opposed to ICET's 4 significant wins. No significant difference was found in the remaining runs. In this setup ACT was able to produce very small trees, sometimes consisting of one node; the accuracy of the learned model was ignored in this setup. ICET, on the contrary, produced, for some of the datasets, larger and more costly trees. DTMC achieved the best results, and outperformed ACT 14 times. The Wilcoxon test also indicates that DTMC is better than ACT and that ACT is better than ICET. Further investigation showed that for a few datasets ACT produced unnecessarily larger trees. We believe that a better tuning of $cf$ would improve ACT in this scenario by making the pruning more aggressive.

At the other extreme, when misclassification costs dominate ($mc = 10000$), the performance of DTMC is worse than ACT and ICET. The t-test indicates that ACT was significantly better than ICET 24 times and significantly worse only 7 times. According to the Wilcoxon test with $\alpha = 5\%$, the difference between ACT and ICET is not significant. Taking $\alpha > 5.05\%$, however, would turn the result in favor of ACT. Observe that DTMC, the winner when $mc = 100$, becomes the worst algorithm when $mc = 10000$. One reason for this phenomenon is that DTMC, as introduced by Ling et al. (2004), does not perform post-pruning, although doing so might improve accuracy in some domains.

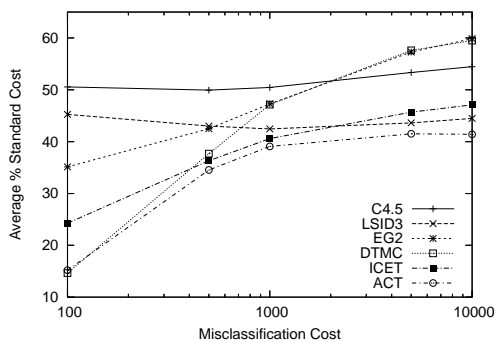

Figure 10: Average normalized cost as a function of misclassification cost

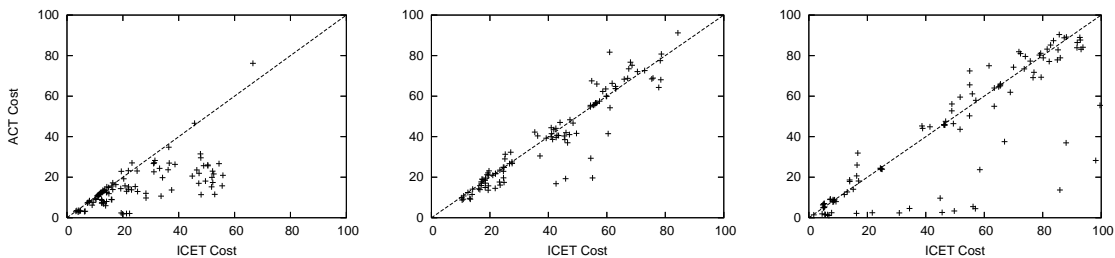

Figure 11: Illustration of the differences in performance between ACT and ICET for $mc = 100, 1000, 10000$ (from left to right). Each point represents a dataset. The $x$-axis represents the cost of ICET while the $y$-axis represents that of ACT. The dashed line indicates equality. Points are below it if ACT performs better and above it if ICET is better.

The above two extremes are less interesting: for the first we could use an algorithm that always outputs a tree of size 1 while for the second we could use cost-insensitive learners. The middle range, where $mc \in \{500, 1000, 5000\}$, requires that the learner carefully balance the two types of cost. In these cases ACT has the lowest average cost and the largest number of t-test wins. Moreover, the Wilcoxon test indicates that it is superior. ICET is the second best method. As reported by Turney (1995), ICET is clearly better than the greedy methods EG2, IDX, and CSID3.

Note that EG2, IDX, and CSID3, which are insensitive to misclassification cost, produced the same trees for all values of $mc$. These trees, however, are judged differently with the change in misclassification cost.

Figure 11 illustrates the differences between ICET and ACT for $mc = 100, 1000, 10000$. Each point represents one of the 105 datasets. The $x$-axis represents the cost of ICET while the $y$-axis represents that of ACT. The dashed line indicates equality. As we can see, the majority of points are below the equality line, indicating that ACT performs better. For $mc = 10000$ we can see that there are points located close to the $x$-axis but with large $x$ value. These points represent the difficult domains, such as XOR, which ICET could not learn but ACT could.

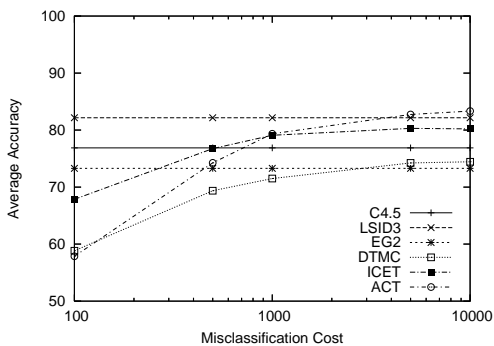

Figure 12: Average accuracy as a function of misclassification cost

### 4.4 Comparing the Accuracy of the Learned Models

When misclassification costs are low, an optimal algorithm would produce a very shallow tree. When misclassification costs are dominant, an optimal algorithm would produce a highly accurate tree. As we can see, ACT's normalized cost increases with the increase in misclassification cost. While it is relatively easy to produce shallow trees, some concepts are not easily learnable and even cost-insensitive algorithms fail to achieve perfect accuracy on them. Hence, as the importance of accuracy increases, the normalized cost increases too because the predictive errors affect it more dramatically.

To learn more about the effect of misclassification costs on accuracy, we compare the accuracy of the built trees for different misclassification costs. Figure 12 shows the results. An important property of DTMC, ICET, and ACT is their ability to compromise on accuracy when needed. They produce inaccurate trees when accuracy is insignificant and much more accurate trees when the penalty for error is high. ACT's flexibility, however, is more noteworthy: from the second least accurate method it becomes the most accurate one.

Interestingly, when accuracy is extremely important, both ICET and ACT achieve even better accuracy than C4.5. The reason is their non-greedy nature. ICET performs an implicit lookahead by reweighting attributes according to their importance. ACT performs lookahead by sampling the space of subtrees before every split. Of the two, the results indicate that ACT's lookahead is more efficient in terms of accuracy. DTMC is less accurate than C4.5. The reason is the different split selection criterion and the different pruning mechanism.

In comparison to our anytime cost insensitive algorithm LSID3, ACT produced less accurate trees when $mc$ was relatively low. When $mc$ was set to 5000, however, ACT achieved comparable accuracy to LSID3 and slightly outperformed it for $mc = 10000$. Statistical tests found the differences between the accuracy of the two algorithms in this case to be insignificant. ACT's small advantage on some of the datasets indicates that, for some problems, expected error is a better heuristic than tree size for maximizing accuracy.

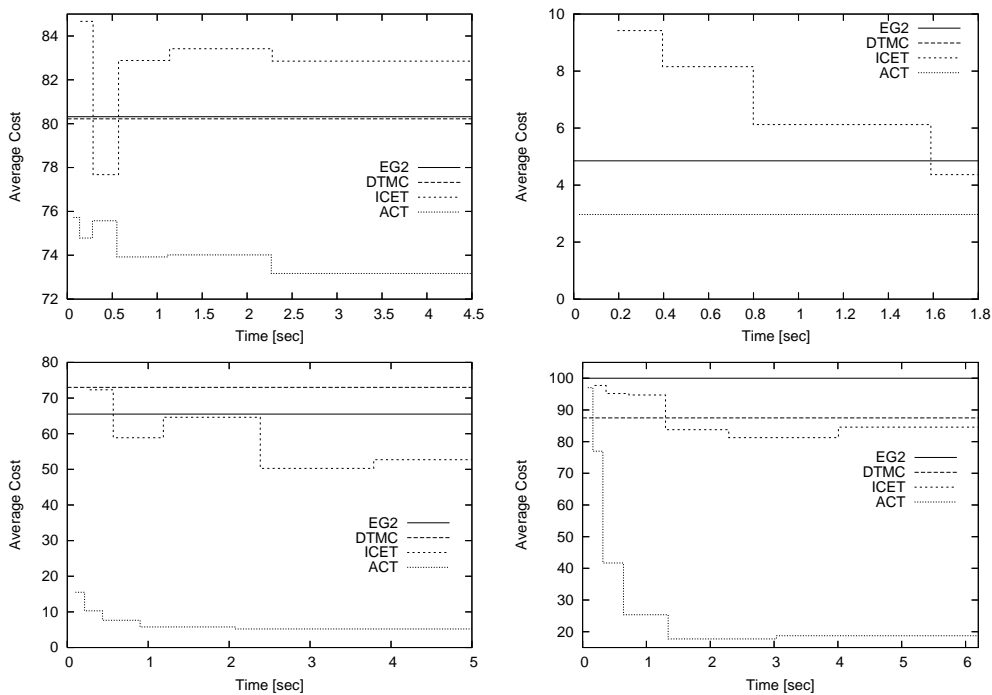

Figure 13: Average normalized cost as a function of time for (from top-left to bottom-right) Breast-cancer-20, Monks1, Multi-XOR, and XOR5

### 4.5 Comparison of Anytime Behavior

Both ICET and ACT, like other typical anytime algorithms, perform better with increased resource allocation. ICET is expected to exploit the extra time by producing more generations and hence better tuning the parameters for the final invocation of EG2. ACT can use the extra time to acquire larger samples and hence achieve better cost estimations.

To examine the anytime behavior of ICET and ACT, we ran them on 4 problems, namely Breast-cancer-20, Monks-1, Multi-XOR, and XOR5, with exponentially increasing time allocation. $mc$ was set to 5000. ICET was run with $2, 4, 8, \ldots$ generations and ACT with a sample size of $1, 2, 4, \ldots$. As in the fixed-time comparison, we used 4 instances for each problem. Figure 13 plots the results averaged over the 4 instances. We also included the results for the greedy methods EG2 and DTMC.

The results show good anytime behavior of both ICET and ACT: generally it is worthwhile to allocate more time. ACT dominates ICET for the four domains and is able to produce less costly trees in shorter time.

One advantage of ACT over ICET is that it is able to consider the context in which an attribute is judged. ICET, on the contrary, reassigns the cost of the attributes globally: an attribute cannot be assigned a high cost in one subtree and a low cost in another. The Multi-XOR dataset exemplifies a concept whose attributes are important only in one sub-concept. The concept is composed of four sub-concepts, each of which relies on different attributes (see Appendix A for further details). As we expected, ACT outperforms ICET significantly because the latter cannot assign context-based costs. Allowing ICET to produce more and more generations (up to 128) does not result in trees comparable to those obtained by ACT.

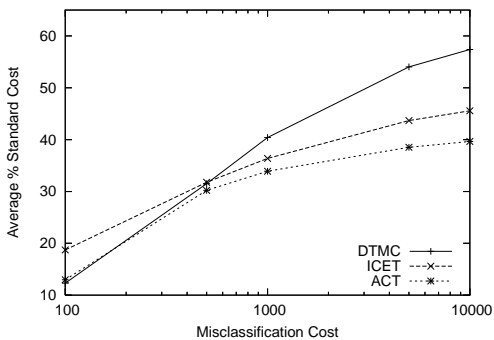

| | DTMC | ICET | ACT |
|---|---|---|---|
| 100 | 12.3 ±1.8 | 18.7 ±2.7 | 13.0 ±2.0 |
| 500 | 31.5 ±3.2 | 31.8 ±3.4 | 30.2 ±3.3 |
| 1000 | 40.4 ±3.9 | 36.4 ±3.9 | 33.9 ±4.0 |
| 5000 | 54.0 ±5.2 | 43.7 ±5.5 | 38.5 ±5.6 |
| 10000 | 57.4 ±5.6 | 45.6 ±5.9 | 39.6 ±6.1 |

Figure 14: Average cost when test costs are assigned randomly

### 4.6 Random Costs

The costs of 100 out of the 105 datasets were assigned using a semi-random mechanism that gives higher costs to informative attributes. To ensure that ACT's success is not due to this particular cost assignment scheme, we repeated the experiments with the costs drawn randomly uniformly from the given cost range $cr$, i.e., $\rho$ was set to 0. Figure 14 shows the results. As we can see, ACT maintains its advantage over the other methods: it dominates them along the scale of $mc$ values.

### 4.7 Nonuniform Misclassification Costs

So far, we have only used uniform misclassification cost matrices, i.e., the cost of each error type was identical. As explained in Section 3, the ACT algorithm can also handle complex misclassification cost matrices where the penalty for one type of error might be higher than the penalty for another type. Our next experiment examines ACT in the nonuniform scenario. Let $FP$ denote the penalty for a false positive and $FN$ the penalty for false negative. When there are more than 2 classes, we split the classes into 2 equal groups according to their order (or randomly if no order exists). We then assign a penalty $FP$ for misclassifying an instance that belongs to the first group and $FN$ for one that belongs to the second group.

To obtain a wide view, we vary the ratio between $FP$ and $FN$ and also examine different absolute values. Table 3 and Figure 15 give the average results. Table 4 lists the number of t-test significant wins each algorithm achieved. It is easy to see that ACT consistently outperforms the other methods.

Table 3: Comparison of C4.5, EG2, DTMC, ACT, and ICET when misclassification costs are nonuniform. *FP* denotes the penalty for a false positive and *FN* the penalty for a false negative. $\gamma$ denotes the basic *mc* unit.

| | | *FP* | $\gamma$ | $\gamma$ | $\gamma$ | $\gamma$ | $2\gamma$ | $4\gamma$ | $8\gamma$ |
| | | *FN* | $8\gamma$ | $4\gamma$ | $2\gamma$ | $\gamma$ | $\gamma$ | $\gamma$ | $\gamma$ |
| $\gamma = 500$ | | C4.5 | 29.2 | 34.2 | 41.3 | 49.9 | 43.6 | 39.0 | 36.3 |
| | | EG2 | 30.1 | 33.0 | 37.2 | 42.5 | 39.3 | 37.5 | 36.8 |
| | | DTCM | 12.4 | 20.3 | 29.8 | 37.7 | 32.5 | 22.9 | 15.8 |
| | | ICET | 23.3 | 27.0 | 31.5 | 36.3 | 34.2 | 31.8 | 29.2 |
| | | ACT | 11.9 | 18.5 | 27.2 | 34.5 | 29.1 | 20.4 | 13.8 |
| $\gamma = 5000$ | | C4.5 | 27.0 | 31.3 | 39.2 | 53.3 | 44.0 | 39.0 | 36.3 |
| | | EG2 | 30.9 | 35.2 | 43.1 | 57.3 | 47.7 | 42.4 | 39.7 |
| | | DTCM | 13.8 | 23.6 | 38.0 | 57.6 | 42.5 | 29.3 | 20.1 |
| | | ICET | 21.4 | 25.6 | 32.7 | 45.7 | 37.4 | 32.8 | 29.8 |
| | | ACT | 12.9 | 19.1 | 28.8 | 41.5 | 31.1 | 22.5 | 14.6 |

Table 4: Comparing DTMC, ACT, and ICET when misclassification costs are nonuniform. For each $FP/FN$ ratio, the columns list the number of t-test significant wins with $\alpha = 5\%$. *FP* denotes the penalty for a false positive and *FN* the penalty for a false negative. $\gamma$ denotes the basic *mc* unit.

| | $\gamma = 500$ | | | | $\gamma = 5000$ | | | |
| $FP/FN$ | DTMC | vs. | ACT | ICET | vs. | ACT | DTMC | vs. | ACT | ICET | vs. | ACT |
| 0.125 | 4 | | 22 | 11 | | 52 | 5 | | 44 | 12 | | 44 |
| 0.25 | 2 | | 31 | 7 | | 49 | 10 | | 49 | 4 | | 36 |
| 0.5 | 10 | | 25 | 7 | | 42 | 16 | | 52 | 10 | | 25 |
| 1 | 9 | | 29 | 5 | | 23 | 7 | | 50 | 15 | | 21 |
| 2 | 5 | | 35 | 1 | | 47 | 5 | | 61 | 9 | | 31 |
| 4 | 3 | | 40 | 0 | | 72 | 4 | | 58 | 0 | | 44 |
| 8 | 1 | | 27 | 4 | | 72 | 0 | | 62 | 0 | | 67 |

Interestingly, the graphs are slightly asymmetric. The reason could be that for some datasets, for example medical ones, it is more difficult to reduce negative errors than positive ones, or vice versa. A similar phenomenon is reported by Turney (1995).

The highest cost for all algorithms is observed when $FP = FN$ because, when the ratio between *FP* and *FN* is extremely large or extremely small, the learner can easily build a small tree whose leaves are labeled with the class that minimizes costs. When misclassification costs are more balanced, however, the learning process becomes much more complicated.

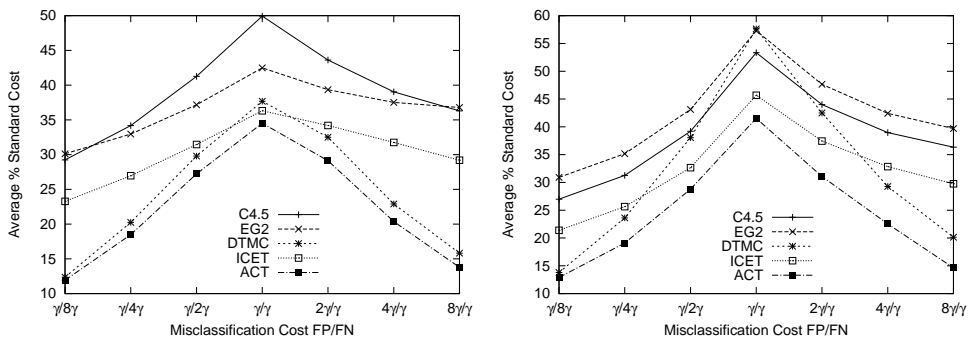

Figure 15: Comparison of C4.5, EG2, DTMC, ACT, and ICET when misclassification costs are nonuniform. The misclassification costs are represented as a pair $(FP/FN)$. $FP$ denotes the penalty for a false positive and $FN$ the penalty for a false negative. $\gamma$ denotes the basic $mc$ unit. The figures plot the average cost as a function of the ratio between $FP$ and $FN$, for $\gamma = 500$ (left) and $\gamma = 5000$ (right).

## 5. Related Work

In addition to the works referred to earlier in this paper, several related works warrant discussion here.

Cost-sensitive trees have been the subject of many research efforts. Several works proposed learning algorithms that consider different misclassification costs (Breiman, Friedman, Olshen, & Stone, 1984; Pazzani, Merz, Murphy, Ali, Hume, & Brunk, 1994; Provost & Buchanan, 1995; Bradford, Kunz, Kohavi, Brunk, & Brodley, 1998; Domingos, 1999; Elkan, 2001; Zadrozny, Langford, & Abe, 2003; Lachiche & Flach, 2003; Abe, Zadrozny, & Langford, 2004; Vadera, 2005; Margineantu, 2005; Zhu, Wu, Khoshgoftaar, & Yong, 2007; Sheng & Ling, 2007). These methods, however, do not consider test costs and hence are appropriate mainly for domains where test costs are not a constraint.

Davis, Ha, Rossbach, Ramadan, and Witchel (2006) presented a greedy cost-sensitive decision tree algorithm for forensic classification: the problem of classifying irreproducible events. In this setup, they assume that all tests that might be used for testing must be acquired and hence charged for before classification.

One way to exploit additional time when searching for a less costly tree is to widen the search space. Bayer-Zubek and Dietterich (2005) formulated the cost-sensitive learning problem as a Markov decision process (MDP), and used a systematic search algorithm based on the AO* heuristic search procedure to solve the MDP. To make AO* efficient, the algorithm uses a two-step lookahead based heuristic. Such limited lookahead is more informed than immediate heuristics but still insufficient for complex domains and might cause the search to go astray (Esmeir & Markovitch, 2007a). The algorithm was shown to output better diagnostic policies than several greedy methods using reasonable resources. An optimal solution, however, could not always be found due to time and memory limits. A nice property of the algorithm is that it can serve as an anytime algorithm by computing the best complete policy found so far. Its anytime behavior, nevertheless, is problematic because policies that are optimal with respect to the train data tend to overfit. As a result, the performance will eventually start to degrade.

Arnt and Zilberstein (2005) tackled the problem of time and cost sensitive classification (TCSC). In TCSC, the utility of labeling an instance depends not only on the correctness of the labeling, but also the amount of time it takes. Therefore the total cost function has an additional component, which reflects the time needed to measure an attribute. Typically, is has a super-linear form: the cost of a quick result is small and fairly constant, but as the waiting time increases, the time cost grows at an increasing rate. The problem is further complicated when a sequence of time-sensitive classification instances is considered, where time spent administering tests for one case can adversely affect the costs of future instances. Arnt and Zilberstein suggest solving these problems by extending the decision theoretic approach introduced by Bayer-Zubek and Dietterich (2005). In our work, we assume that the time it takes to administer a test is incorporated into its cost. In the future, we intend to extend our framework to support time-sensitive classification, both for individual cases and for sequences.

Fan, Lee, Stolfo, and Miller (2000) studied the problem of cost-sensitive intrusion detection systems (IDS). The goal is to maximize security while minimizing costs. Each prediction (action) has a cost. Features are categorized into three cost levels according to amount of information needed to compute their values. To reduce the cost of an IDS, high cost rules are considered only when the predictions of low cost rules are not sufficiently accurate.

Costs are also involved in the learning phase, during example acquisition and during model learning. The problem of budgeted learning has been studied by Lizotte, Madani, and Greiner (2003). There is a cost associated with obtaining each attribute value of a training example, and the task is to determine what attributes to test given a budget.

A related problem is active feature-value acquisition. In this setup one tries to reduce the cost of improving accuracy by identifying highly informative instances. Melville, Saar-Tsechansky, Provost, and Mooney (2004) introduced an approach in which instances are selected for acquisition based on the accuracy of the current model and its confidence in the prediction.

Greiner, Grove, and Roth (2002) were pioneers in studying classifiers that actively decide what tests to administer. They defined an *active classifier* as a classifier that given a partially specified instance, returns either a class label or a strategy that specifies which test should be performed next. Greiner et al. also analyzed the theoretical aspects of learning optimal active classifiers using a variant of the probably-approximately-correct (PAC) model. They showed that the task of learning optimal cost-sensitive active classifiers is often intractable. However, this task is shown to be achievable when the active classifier is allowed to perform only (at most) a constant number of tests, where the limit is provided before learning. For this setup they proposed taking a dynamic programming approach to build trees of at most depth $d$.

Our setup assumed that we are charged for acquiring each of the feature values of the test cases. The term test strategy (Sheng, Ling, & Yang, 2005) describes the process of feature values acquisition: which values to query for and in what order. Several test strategies have been studied, including sequential, single batch and multiple batch (Sheng et al., 2006), each of which corresponds to a different diagnosis policy. These strategies are orthogonal to our work because they assume a given decision tree.

Bilgic and Getoor (2007) tackled the problem of feature subset selection when costs are involved. The objective is to minimize the sum of the information acquisition cost and the misclassification cost. Unlike greedy approaches that compute the value of features one at a time, they used a novel data structure called the value of information lattice (VOILA), which exploits dependencies between missing features and makes it possible to share information value computations between different feature subsets possible. VIOLA was shown empirically to achieve dramatic cost improvements without the prohibitive computational costs of comprehensive search.

## 6. Conclusions

Machine learning techniques are increasingly being used to produce a wide range of classifiers for complex real-world applications that involve nonuniform testing and misclassification costs. The increasing complexity of these applications poses a real challenge to resource management during learning and classification. In this work we introduced a novel framework for operating in such complex environments. Our framework has four major advantages:

- It uses a non-greedy approach to build a decision tree and therefore is able to overcome local minima problems.

- It evaluates entire trees during the search; thus, it can be adjusted to any cost scheme that is defined over trees.

- It exhibits good anytime behavior and allows learning speed to be traded for classification costs. In many applications we are willing to allocate more time than we would allocate to greedy methods. Our proposed framework can exploit such extra resources.

- The sampling process can easily be parallelized and the method benefit from distributed computer power.

To evaluate ACT we have designed an extensive set of experiments with a wide range of costs. Since there are only a few publicly available cost-oriented datasets, we designed a parametric scheme that automatically assigns costs for existing datasets. The experimental results show that ACT is superior to ICET and DTMC, existing cost-sensitive algorithms that attempt to minimize test costs and misclassification costs simultaneously. Significance tests found the differences to be statistically strong. ACT also exhibited good anytime behavior: with the increase in time allocation, the cost of the learned models decreased.

ACT is a contract anytime algorithm that requires its sample size to be predetermined. In the future we intend to convert ACT into an interruptible anytime algorithm by adopting the IIDT general framework (Esmeir & Markovitch, 2007a). In addition, we plan to apply monitoring techniques (Hansen & Zilberstein, 2001) for optimal scheduling of ACT and to examine other strategies for evaluating subtrees.

Table 5: Characteristics of the datasets used

| Dataset | Instances | Attributes Nominal (binary) | Numeric | Max attribute domain | Classes |
|---|---|---|---|---|---|
| Breast Cancer | 277 | 9 (3) | 0 | 13 | 2 |
| Bupa | 345 | 0 (0) | 5 | - | 2 |
| Car | 1728 | 6 (0) | 0 | 4 | 4 |
| Flare | 323 | 10 (5) | 0 | 7 | 4 |
| Glass | 214 | 0 (0) | 9 | - | 7 |
| Heart | 296 | 8(4) | 5 | 4 | 2 |
| Hepatitis | 154 | 13(13) | 6 | 2 | 2 |
| Iris | 150 | 0 (0) | 4 | - | 3 |
| KRK | 28056 | 6(0) | 0 | 8 | 17 |
| Monks-1 | 124+432 | 6 (2) | 0 | 4 | 2 |
| Monks-2 | 169+432 | 6 (2) | 0 | 4 | 2 |
| Monks-3 | 122+432 | 6 (2) | 0 | 4 | 2 |
| Multiplexer-20 | 615 | 20 (20) | 0 | 2 | 2 |
| Multi-XOR | 200 | 11 (11) | 0 | 2 | 2 |
| Multi-AND-OR | 200 | 11 (11) | 0 | 2 | 2 |
| Nursery | 8703 | 8(8) | 0 | 5 | 5 |
| Pima | 768 | 0(0) | 8 | - | 2 |
| TAE | 151 | 4(1) | 1 | 26 | 3 |
| Tic-Tac-Toe | 958 | 9 (0) | 0 | 3 | 2 |
| Titanic | 2201 | 3(2) | 0 | 4 | 2 |
| Thyroid | 3772 | 15(15) | 5 | 2 | 3 |
| Voting | 232 | 16 (16) | 0 | 2 | 2 |
| Wine | 178 | 0 (0) | 13 | - | 3 |
| XOR 3D | 200 | 0 (0) | 6 | - | 2 |
| XOR-5 | 200 | 10 (10) | 0 | 2 | 2 |

## Acknowledgments

This work was partially supported by funding from the EC-sponsored MUSCLE Network of Excellence (FP6-507752).

## Appendix A. Datasets

Table 5 lists the characteristics of the 25 datasets we used. Below we give a more detailed description of the non-UCI datasets used in our experiments:

1. *Multiplexer:* The multiplexer task was used by several researchers for evaluating classifiers (e.g., Quinlan, 1993). An instance is a series of bits of length $a + 2^a$, where $a$ is a positive integer. The first $a$ bits represent an index into the remaining bits and the label of the instance is the value of the indexed bit. In our experiments we considered the 20-Multiplexer ($a = 4$). The dataset contains 500 randomly drawn instances.

2. *Boolean XOR:* Parity-like functions are known to be problematic for many learning algorithms. However, they naturally arise in real-world data, such as the Drosophila survival concept (Page & Ray, 2003). We considered XOR of five variables with five additional irrelevant attributes.

3. *Numeric XOR:* A XOR based numeric dataset that has been used to evaluate learning algorithms (e.g., Baram, El-Yaniv, & Luz, 2003). Each example consists of values for $x$ and $y$ coordinates. The example is labeled 1 if the product of $x$ and $y$ is positive, and $-1$ otherwise. We generalized this domain for three dimensions and added irrelevant variables to make the concept harder.

4. *Multi-XOR / Multi-AND-OR:* These concepts are defined over 11 binary attributes. In both cases the target concept is composed of several subconcepts, where the first two attributes determines which of them is considered. The other 10 attributes are used to form the subconcepts. In the Multi-XOR dataset, each subconcept is an XOR, and in the Multi-AND-OR dataset, each subconcept is either AND or OR.